\documentclass{article}

\usepackage{amssymb}
\usepackage{PRIMEarxiv}

\usepackage[utf8]{inputenc} 
\usepackage[T1]{fontenc}    
\usepackage{hyperref}       
\usepackage{url}            
\usepackage{booktabs}       
\usepackage{amsfonts}       
\usepackage{nicefrac}       
\usepackage{microtype}      
\usepackage{lipsum}
\usepackage{fancyhdr}       
\usepackage{graphicx}       
\usepackage{multirow}
\usepackage{subfigure}
\graphicspath{{media/}}     

\title{TraClets: Harnessing the power of computer vision for trajectory classification
}

\author{
  Ioannis Kontopoulos, Antonios Makris, Konstantinos Tserpes \\
  Department of Informatics and Telematics \\
  Harokopio University \\
  Athens, Greece\\
  \texttt{\{kontopoulos, amakris, tserpes\}@hua.gr} \\
   \And
  Vania Bogorny \\
  Programa de P\'{o}s-Gradua\c{c}\~{a}o em Ci\^{e}ncia da Computa\c{c}\~{a}o \\
  Universidade Federal de Santa Catarina (UFSC) \\
  Florian\'{o}polis, Brazil\\
  \texttt{vania.bogorny@ufsc.br} \\
}

\begin{document}
\maketitle

\begin{abstract}
Due to the advent of new mobile devices and tracking sensors in recent years, huge amounts of data are being produced every day. Therefore, novel methodologies need to emerge that dive through this vast sea of information and generate insights and meaningful information. To this end, researchers have developed several trajectory classification algorithms over the years that are able to annotate tracking data. Similarly, in this research, a novel methodology is presented that exploits image representations of trajectories, called TraClets, in order to classify trajectories in an intuitive humans way, through computer vision techniques. Several real-world datasets are used to evaluate the proposed approach and compare its classification performance to other state-of-the-art trajectory classification algorithms. Experimental results demonstrate that TraClets achieves a classification performance that is comparable to, or in most cases, better than the state-of-the-art, acting as a universal, high-accuracy approach for trajectory classification.
\end{abstract}


\section{Introduction}

The sudden increase in mobile devices and tracking sensors has led to the creation of an abundance of data. These kind of data have shifted both researchers' and industries' attention alike towards algorithms and methodologies able to extract knowledge from a large pool of spatio-temporal data (e.g. trajectory classification). Trajectory classification is an important mining task as it can uncover information regarding potential animal migration patterns \cite{DBLP:journals/gis/DamianiIFHC16,https://doi.org/10.1111/2041-210X.12453}, the type and strength level of hurricanes \cite{DBLP:journals/pvldb/LeeHLG08} and illegal vessel activities at sea \cite{DBLP:conf/debs/Kontopoulos0TZ20}, thus improving overall surveillance and safety in various domains.

Several approaches for trajectory classification \cite{silva2019,DBLP:conf/debs/Kontopoulos0TZ20} try to exploit global features such as average speed, acceleration or the standard deviation of them. Other approaches try to apply trajectory partitioning \cite{DBLP:journals/pvldb/LeeHLG08} or find relevant sub-trajectories \cite{ferrero2018movelets} in an attempt to identify more discriminative features. The context of the analysis is typically the physical world and the geography. Position, speed and acceleration are the basic features in a possibly multi-dimensional space. However, experts rely heavily on the visualization of trajectories to manually classify trajectories that are of some significance. This provides an intuition to move the analysis in a different domain, leveraging computer vision approaches on classification. In computer vision, the most commonly used techniques include Convolutional Neural Networks (CNNs) \cite{Rawat2017,Zhong2011,Wu2015}. Each layer of a CNN identifies a different feature of the image, including but not limited to shape and color. As an attempt to increase the performance of CNNs, deep learning emerged \cite{Zhong2011,Wu2015}, increasing the complexity of the networks. One of the most common goals of such networks is to classify a set of images to a predefined set of labels of interest.

Our proposed methodology builds upon previous works \cite{kontopoulos2021computer,DBLP:journals/ijgi/KontopoulosMT21} that represent trajectories of moving objects as images and accurately classify them, by incorporating the notion of acceleration to the images. Furthermore, in this work we compare our proposed methodology to four state-of-the-art trajectory classifiers  over three well-known datasets. The novelty of our approach lies in the use of a computer vision methodology for the classification of moving objects' trajectories. Specifically, the proposed approach leverages an image classification approach by visually representing trajectories as images, called TraClets. The use of TraClets for the problem at hand and its main contributions are:

\begin{itemize}
    \item patterns formed by the trajectories of moving objects tend to be visually distinct \cite{kontopoulos2021computer}. This visual distinctiveness leads to an increased trajectory classification performance;
    \item image classification works for  trajectory classification even when data are not transmitted at fixed intervals (e.g. hourly). This is in contrast to time-series methodologies \cite{DBLP:journals/datamine/FawazFWIM19} that are inherently unsuitable for such tasks and require data points at fixed time points;
    \item most trajectory classification approaches found in the literature \cite{da2019survey} require a pre-processing step that involves the understanding and analysis of data and the selection of features suitable only for the moving objects' trajectories to be classified. This means that features selected for a certain trajectory (e.g. cars) cannot be applied to other patterns as well (e.g. vessels) \cite{Souza2016}. A computer vision approach for trajectory classification skips entirely the aforementioned pre-processing step; the same technique for classifying an image (e.g. CNNs) can be applied for the classification of other trajectories since they are transformed into images. Therefore, a computer vision approach for trajectory classification yields a promising universal approach for the classification of moving objects' trajectories;
    \item the partitioning of trajectories often constitutes a first step when dealing with trajectory classification \cite{lee2007trajectory,DBLP:journals/pvldb/LeeHLG08}. Such approaches often require input parameters which are hard to determine and that eventually have a significant impact on the partitioning results. As our method entirely skips this step, as stated above, we have eliminated the need of arbitrary or empirical user-defined parameters, making our approach scalable and robust.
\end{itemize}

The rest of the paper is organized as follows. Section \ref{rel_work} serves as a literature review in the field of trajectory classification. Section \ref{methodology} describes in detail the TraClets and the way they are used in the classification process while Section \ref{exp_evaluation} evaluates the proposed approach in terms of classification performance and compares the TraClets with other techniques. Finally, Section \ref{conclusion} summarizes the merits of our work and highlights some perspectives that require further attention in the future.

\section{Related Work}
\label{rel_work}

Over the years, several techniques for trajectory classification have been proposed, such as transportation mode prediction of mobility objects,  animal specie identification, hurricane level determination, etc. Trajectory classification methods extract different features from the spatiotemporal properties of trajectories, such as speed, acceleration and direction change, to use as input for trajectory classifiers. As referred in \cite{da2019survey}, the difference between the classification techniques is focused on the type of trajectory features extracted for creating the classification model. Effective trajectory classification requires generating a set of features that discriminate the class.
 
One of the first trajectory classification methods was TRACLASS \cite{lee2008traclass}. The framework generated a hierarchy of features by combining two types of clustering: i) region-based which discovers regions of trajectories that belong on one class and ii) trajectory-based which discovers sub-trajectories that present common moving patterns of each class. It is based on dividing the space in grids, and by reducing the grid sizes until the trajectories inside the grid belong to the same class. If the trajectories inside a grid cell are from the same class then it is selected as a feature, otherwise its size is reduced until the lowest possible size, given by a threshold. If the lowest threshold size is reached and the trajectories inside the grid are not from the same class, then the trajectories are split by direction change, and these subtrajectories are clustered using a density-based clustering algorithm. The grid cells and subtrajectory clusters are used as the features to feed the classification model, by identifying whether the trajectory is part of that features or not.

The works of Bolbol \cite{bolbol2012inferring}, Soleymani \cite{soleymani2014integrating}, Dabiri \cite{dabiri2018inferring}, and \cite{vicenzi2020} extract local features from trajectories, i.e., attributes from trajectory parts. More specifically, Bolbol \cite{bolbol2012inferring} segments the trajectories in a pre-defined number of subtrajectories and uses a sliding window to cover a certain number of subtrajectories. The proposed framework is based on Support Vector Machines (SVMs) classification and extracts features as the average acceleration and average speed of the trajectory. Soleymani \cite{soleymani2014integrating} computes and analyzes features in both spatial and temporal domains. The method segments the trajectories by using two types of grids. The first grid splits the trajectories based on their spatial location, and the technique extracts the time duration of the subtrajectories inside each grid cell. The second grid divides the trajectory by using a time window, and the technique calculates features as the standard deviation of speed and maximum turning angle from each subtrajectory inside a grid cell. Dabiri \cite{dabiri2018inferring} presented a Convolutional Neural Network architecture for trajectory mode classification. Four features from sequential trajectory points (speed, acceleration, direction change and stop rate) are calculated. Then, trajectories are represented by a vector of four dimensions, one for each feature. This vector is fed into the CNN to estimate the transportation mode. Vicenzi in \cite{vicenzi2020} extracts local features from each trajectory point and selects the most frequent ones to use as input to a classifier. However, the process of finding the most discriminant dimension must be done manually by the user. 

The works of Zheng \cite{zheng2008understanding}, Sharma  \cite{sharma2010nearest} and Junior  \cite{junior2017analytic} extract global features, i.e., features from the entire trajectory.
Zheng \cite{zheng2008understanding} propose a supervised learning approach for transportation mode classification. For each trajectory are extracted features such as length, maximum speed, acceleration, average expectation and variance of speed, heading change rate, stop rate and velocity change rate. Subsequently, these features are used for training a Decision Tree-based model in order to perform predictions. Junior \cite{junior2017analytic} presented an active learning approach called ANALYTiC, which enabled semantic annotation on the learning set. The proposed approach extracts the maximum, minimum and average speed, direction change and traveled distance for training an Active Learning model. Sharma \cite{sharma2010nearest} uses the Nearest Neighbour Trajectory Classification (NNTC), where a trajectory is assigned to the same class as its neighbour, i.e., the closest trajectory. For data enrichment, an initial pre-processing is performed which adds additional features such as the time of the next position in the sequence, the time interval and the distance in space to the next position, the speed, direction, acceleration and direction change.

A segmentation and feature extraction method for trajectories which considers local and global features was presented in \cite{dodge2009revealing}. The  method calculates features every two consecutive trajectory points and then the trajectories are represented as sequences of each feature. Local features were extracted from sub-trajectories with the same characteristics (e.g. same speed, same acceleration, etc.) and global features were statistics of the entire trajectory as the minimum speed, maximum speed, average speed, minimum acceleration, etc.
Xiao \cite{xiao2017identifying} and Etemad \cite{etemad2018predicting} extract several statistics from trajectory points, as the percentiles, interquatile range, skewness, coefficient of variation and kurtosis from the speed, acceleration, and heading change.
Patel in \cite{patel2012incorporating} extended the work of Lee  \cite{lee2008traclass} by considering the temporal dimension in the grid cells, where the local features were the grids, and the global features were, for example, the total duration and the traveled distance of a trajectory.
A trajectory classification method called Movelets, which is based on shapelet analysis is presented in \cite{ferrero2018movelets}. The Movelets technique extracts relevant sub-trajectories called movelets, in order to generate local features of each trajectory and compares the distance of each sub-trajectory to all trajectories in the dataset. Movelets is able to consider all trajectory dimensions as space, time and semantic information. An extension of this method is proposed in \cite{ferrero2020mastermovelets} to choose the best dimension combination, but as movelets, it still has a heavy preprocessing step that needs to explore all sub-trajectory sizes and dimension combinations to achieve a high accuracy, while the approach proposed in this paper skips this expensive step.

In addition, neural networks have been widely used for trajectory classification. Jiang et al. \cite{jiang2017trajectorynet} employed Recurrent Neural Networks (RNNs) for point-based trajectory classification, by utilizing embedding of GPS data to map the original low-dimensional and heterogeneous feature space into distributed vector representations. The feature space was enriched with segment-based information, and maxout activations were employed by RNNs for an increased performance. Zhang et al. \cite{zhang2019classifying} employed a deep multi-scale learning model to model grid data under different space and time granularities, thus capturing the impact of space and time on the classification results. An attention dense module was designed by combining an attention mechanism, which was able to select major features, and the DenseNet model, which was able to enhance the propagation of local and spatial features throughout the network. Furthermore, Jiang et al. \cite{jiang2017improving} employed RNNs and specifically a novel partition-wise Gated Recurrent Unit (pGRU) architecture for point-based trajectory classification on detecting fishing activities. The proposed method maps low-dimensional features into another space with the use of a partition-wise activation function applied before the linear transformation and then receives different parameters for distinct partitions, to model them jointly in a nonlinear hierarchical deep structure. Thus, the proposed method is less sensitive to the quality of partitions and achieves better predictive results. In \cite{Petry2020}, Petry proposed to use embeddings over the trajectory dimensions of each point, and for the spatial dimension the geohash is used in the embedding. Finally, Makris et al. \cite{makris2021semi} proposed a semi-supervised convolutional auto-encoder (CAE) model for vessel trajectory classification. While classic auto-encoders are often limited to learning low-level structures of an image such as lines and edges, convolutional auto-encoders can cope with multidimensional images. The main advantage of CAEs over traditional supervised CNNs is their ability to achieve high classification performance in many applications.

\section{Methodology}
\label{methodology}

This section describes the creation of  TraClets from the trajectories and the way these TraClets are employed in the classification process via a Convolutional Neural Network.

\subsection{TraClets}

In this section we present TraClets, the image representation of  trajectories. These representations are indicative of the mobility patterns of the moving objects. TraClets need to efficiently visualize and capture three key features that characterize the trajectory patterns of moving objects: i) the shape of the trajectory which indicates the way the object moves in space taking into account the heading, ii) the speed, and iii) the acceleration that indicates how fast the object moves in space.

\subsubsection{Trajectory shape}

Trajectories of the similar moving objects (e.g. cars) tend to form similar patterns in terms of the way the object moves. However, the distance each moving object travels through space is different (e.g. a car in a highway travels greater distances compared to a car in an urban area). Therefore, the bounding box or the surveillance area in which the object moves needs to be normalized. To this end, the total distance of both the $x$ and the $y$ axis in which the object moves must be defined first. For this reason, both the total horizontal distance (Eq. \ref{xdistance}) and the total vertical distance the object has travelled are calculated (Eq. \ref{ydistance}) based on the minimum and maximum longitudes and latitudes respectively. The total horizontal distance is defined as: 

\begin{equation}
    d_x = longitude_{max} - longitude_{min}\label{xdistance}
\end{equation}

and the total vertical distance the object has travelled is defined as:

\begin{equation}
    d_y = latitude_{max} - latitude_{min}\label{ydistance}
\end{equation}

Then, the distance that each position $m$ has travelled from the minimum longitude and latitude can be calculated from the equations \ref{xposdistance} and \ref{yposdistance} respectively as follows:

\begin{equation}
    d(m_x) = longitude_{m} - longitude_{min}\label{xposdistance}
\end{equation}

and

\begin{equation}
    d(m_y) = latitude_{m} - latitude_{min}\label{yposdistance}
\end{equation}

From equations \ref{xdistance},\ref{xposdistance} and \ref{ydistance},\ref{yposdistance}, the percentage of the total distance each position $m$ has travelled so far can be calculated from the minimum coordinate in both $x$ and $y$ axes:

\begin{equation}
    normalized(m_x) = d(m_x) \div d_x\label{xpercentage}
\end{equation}

and

\begin{equation}
    normalized(m_y) = d(m_y) \div d_y\label{ypercentage}
\end{equation}

Given a predefined image size of $N \times N$, the exact position of $m$ inside an image can be calculated as follows:

\begin{equation}
    pixel_x = normalized(m_x) \times N\label{xpixel}
\end{equation}

and

\begin{equation}
    pixel_y = normalized(m_y) \times N\label{ypixel}
\end{equation}

\begin{figure}[ht]
\centering     
\subfigure[Positions of the trajectory.]{\label{fig:space}\includegraphics[scale=0.25]{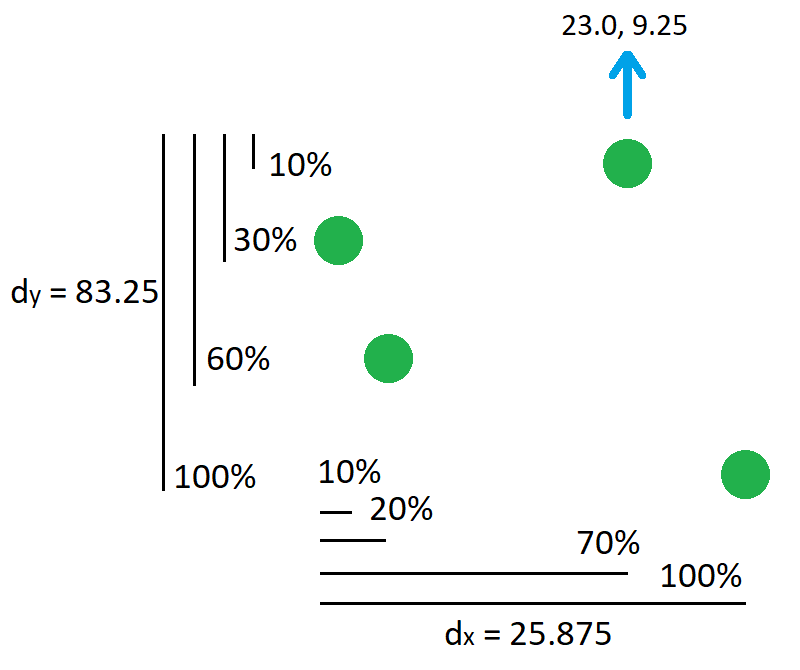}}
\subfigure[Positions placed in a $10 \times 10$ raster.]{\label{fig:pixel_space}\includegraphics[scale=0.25]{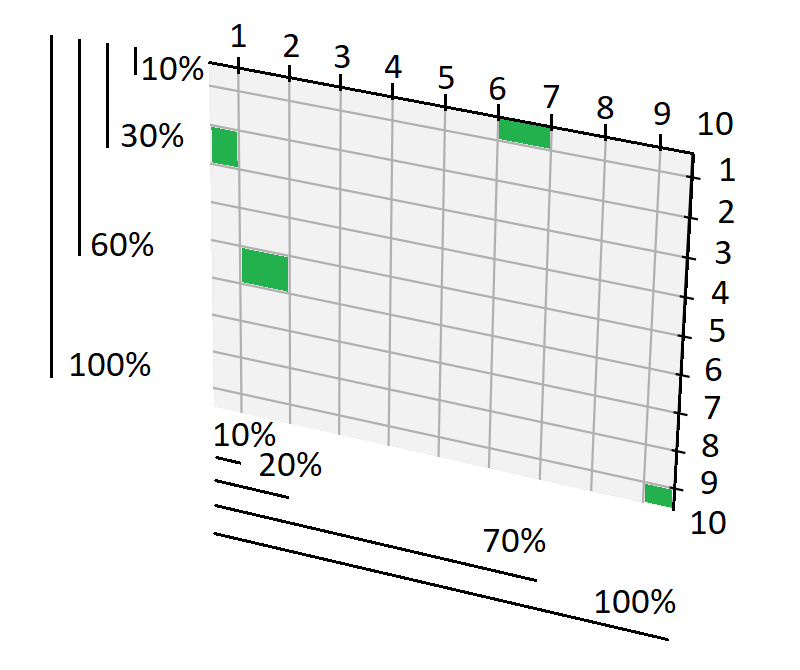}}
\caption{An example of space normalization.}
\label{fig:normalization}
\end{figure}

Therefore, each position is placed inside a normalized bounding box or a surveillance space of size $N \times N$ that is essentially an image representation. Figure \ref{fig:normalization} demonstrates an example of the surveillance space normalization for $N = 10$. Each green circle corresponds to a position. The total horizontal distance is $\mathit{d_x = 28.75 - 2.875 = 25.875}$ while the total vertical distance is $\mathit{d_y = 92.5 - 9.25 = 83.25}$. Based on equation \ref{xposdistance} and \ref{yposdistance}, $d(m_x)$ of the upper right position ($longitude = 23.0, latitude = 9.25$) is equal to $20.125$ and $(m_y)$ is equal to $0$. According to equations \ref{xpercentage} and \ref{ypercentage}, $normalized(m_x)$ is equal to $0.7$ and $normalized(m_y)$ is equal to $0$. Multiplying each normalized value by $N = 10$ results in $pixel_x = 7$ and $pixel_y = 0$. In order to fit boundary trajectory positions into the normalized surveillance space, pixel values smaller than $N$ are transformed into 1 and pixel values larger than $N$ are transformed into $N$. Therefore, $pixel_y$ is transformed into $1$ and the final pixel position inside the $10 \times 10$ image is $x=7$ and $y=1$ as shown in Figure \ref{fig:normalization}.

\subsubsection{Trajectory speed}
\label{speed_rep}

To capture the changes in speed, previous research conducted on the classification of vessel trajectories \cite{kontopoulos2021computer,DBLP:journals/ijgi/KontopoulosMT21} is taken into consideration, in which the maximum speed $max(speed)$ of the vessels was segmented to $11$ equally sized increments with each speed increment corresponding to a different RGB color value in the final image. Let $S=[0,max(speed)]$ be the possible speed values at which the moving objects move. A speed increment $incr(speed)$ is defined as:

\begin{equation}
    incr(speed) = max(speed) \div 11\label{speed_incr}
\end{equation}

resulting in an arithmetic sequence such that:

\begin{equation}
    \alpha_i = \alpha_1 + (i-1) \times incr(speed)\label{speed_sequence}
\end{equation}

where $\alpha_i$ is the $i^{th}$ term of the sequence, $\alpha_1$ is the initial term of the sequence (in our case $0~m/s$) and $incr(speed)$ is the common difference of successive members. Each pixel $p(pixel_x,pixel_y)$ that corresponds to a position $m$ is colored with a different color value that corresponds to the $i^{th}$ term of the sequence depending on the speed the position has at the moment. Moreover, pixels that do not contain any positions are colored white.

\subsubsection{Trajectory acceleration}
\label{accel_rep}

\begin{figure}
  \centering
  \includegraphics[width=0.3\linewidth]{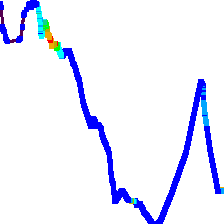}
  \caption{An example of a TraClet.}
  \label{fig:traclet}
\end{figure}

To represent the acceleration of the trajectory, a straight line between each temporally consecutive $pixel(pixel_x,pixel_y)$ or position $m$ is drawn initially using the Bresenham's line algorithm \cite{DBLP:journals/jcp/Gaol13}, in order to make the pattern created by each moving object more distinctive. The Bresenham's line algorithm is a line-drawing algorithm that generates points that form a close approximation to a straight line between two points of an $N$-dimensional raster. Then, similar to Section \ref{speed_rep}, the acceleration is segmented to $11$ equally sized increments and each line is colored based on the absolute value of the acceleration and the $i^{th}$ term of the acceleration sequence that the acceleration value falls into. The acceleration of the line is defined as the acceleration between the starting and ending position $m$ of the respective line. The final result is an image, called TraClet, that depicts the movement, the speed and the acceleration of the trajectory. An example of a TraClet is illustrated in Figure \ref{fig:traclet}.

\subsection{Deep learning for trajectory classification}
\label{DL_for_TrajCla}

This section describes the deep learning approach employed for mobility data classification from trajectory images, based on fine-tuning and Convolutional Neural Networks.

CNNs are able to learn deep features of trajectories such as shape and color (e.g., speed, acceleration) automatically, thus eliminating hand-created feature extraction. Furthermore, they are able to capture more local spatial correlations, as neurons in the CNN receive signals from other neurons in the local area in the preceding layer \cite{chen2019mobility}. In addition, the weight-sharing characteristic in the connection of adjacent layers can significantly reduce the number of variables. The main disadvantage of image classification through deep learning is that these methods require a large amount of data in order to perform accurate feature extraction and classification. In order to overcome this limitation, transfer learning was adopted. Transfer learning is a common and effective method, which aims at training a network with fewer samples, as the knowledge extracted by a pre-trained model is then reused and applied to the given task of interest.
The intuition behind transfer learning is that generic features learned on a general large dataset, can be shared among seemingly disparate datasets. The learned features can be used to solve a different, but related task \cite{pan2009survey}.

\begin{figure}
 \centering
 \includegraphics[width=0.94\linewidth]{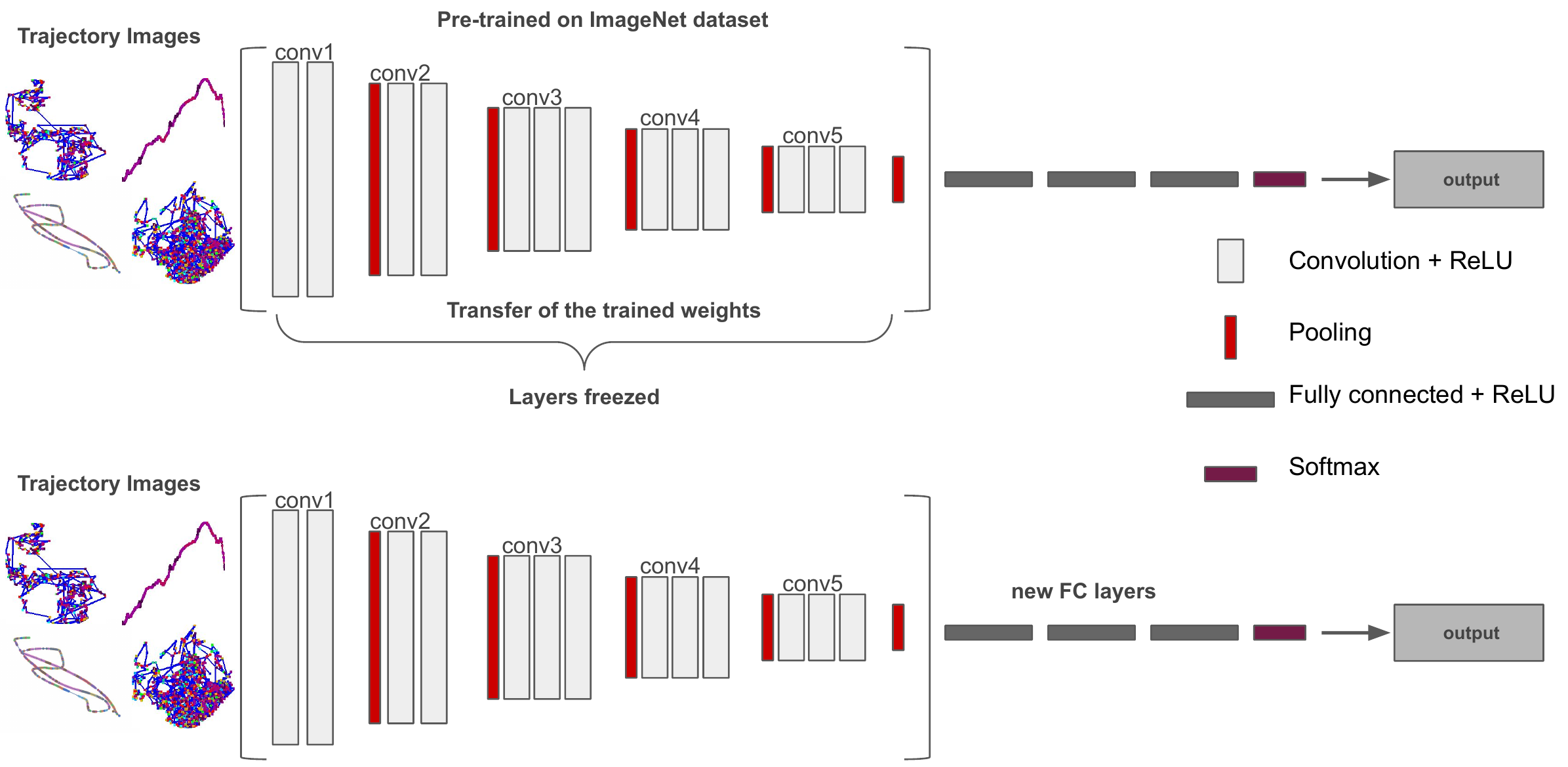}
 \caption{Fine-tuning on the VGG16 network architecture.}
 \label{trlearn_architect}
\end{figure}

Fine-tuning was employed as the standard method for transfer learning. The weights were pre-trained on the ImageNet \cite{deng2009imagenet} dataset for the examined CNN, namely VGG16.
Figure \ref{trlearn_architect} illustrates the fine-tuning process on the VGG16 network.
The VGG16 model consists of $13$ convolutional and $3$ fully connected ($FC$) layers. The final set of layers which contain the $FC$ layers along with the activation function is called the ``head''. The network was instantiated with weights pre-trained on ImageNet, as shown at the top of the figure. Afterwards, the $FC$ layers were truncated, and the final $pooling$ layer was treated as a feature extractor. The body of the network, i.e., the weights of the convolution layers of the pre-trained network, were freezed such that only the $FC$ head layer was trained. This was because the convolution layers had already learned discriminative filters and captured universal features like curves and edges; thus, these weights had to remain intact. Finally, the truncated layers were replaced by a new $FC$ head layer, which was randomly initialized and placed on top of the original architecture (bottom of the figure). In other words, the $FC$ head layer was randomly initialized from scratch and focused on learning dataset-specific features.

Trajectory data were transformed into images and labeled based on concrete mobility patterns' annotation into several classes based on the dataset. Animals dataset for example, contains three classes which represent three animal species, namely elk, mule deer and cattle. 
These images, i.e. TraClets, were fed as input into the deep learning model for training.  
Input images were scaled to the fixed size of $224 \times 224$ pixels. Training was conducted for $100$ epochs with a learning rate of $1e - 3$ and a batch size of $16$. 
The output of the convolution layers was activated by the non-linear activation function called the Rectified Linear Unit (ReLU), due to its reduced likelihood of vanishing gradients and its efficient computation. 
After each convolution layer, the pooling layer was introduced to carry out downsampling operations, which reduced the in-plane dimensionality of the feature maps. Downsample operations after convolutional layers introduce a translation invariance to small shifts and distortions and decrease the number of subsequent learnable parameters. Average pooling \cite{lin2013network} was employed as the pooling strategy and the number of learnable parameters was reduced, preventing overfitting. 
The output feature maps of the final layer were flattened and connected to the fully connected layers, in which every input was connected to every output by a learnable weight. The output layer generated a probability distribution over the classification labels, by resorting the softmax function. As we are facing multi-class classification, the ``categorical\_crossentropy'' is utilized as the loss function which  compares the distribution of the predictions with the true distribution. 

Adam optimization method \cite{kingma2014adam} was used to find the minimum of the objective (error) function making use of its gradient.
Furthermore, a dropout layer \cite{hinton2012improving} of $0.2$ was applied, which means that 20\% of neurons were randomly set to zero during each training epoch, thus avoiding overfitting on the training dataset. Dropout is one of the most popular regularization techniques, which forces the weights in the network to receive only small values, making the distribution of weight values more regular. As a result, this technique is able to reduce overfitting on small training examples \cite{hawkins2004problem}.
In addition, in order to prevent model overfitting, data augmentation was performed during training, leveraging several multi-processing techniques. Specifically, the transformations employed included random rotation of the images (the maximum rotation angle was 30 degrees), horizontal flips, shearing, zooming, cropping, and small random noise perturbations. Data augmentation improved the generalization and enhanced the learning capability of the model. Table \ref{tab:tuning_parameters_CNN} summarizes the parameters employed by the model.

\begin{table}
\centering
\begin{tabular}{ll}
\hline\noalign{\smallskip}
Parameter	& Value\\
\noalign{\smallskip}\hline\noalign{\smallskip}
 Epochs & $100$ \\
 Learning rate & \(1e - 3\) \\
 Batch size & $16$ \\
 Activation function & $ReLU$ \\
 Pooling & $Average2D$ \\
 Optimizer & $Adam$ \\
 Output activation function & $Softmax$ \\
 Loss function & $categorical\_crossentropy$ \\
 Dropout probability & $0.2$ \\
\noalign{\smallskip}\hline
\end{tabular}
\caption{Configuration of the CNN architecture\label{tab:tuning_parameters_CNN}}
\end{table}

We have implemented an open-source deep learning toolkit called d-Look\footnote{\url{https://github.com/AntonisMakris/d-LOOK}} under the MIT Licence, which provides an automated way to execute various supervised deep learning Keras\footnote{\url{https://keras.io/}} models. In this research work, we have utilized only one model provided by d-Look, VGG16, as it is proven to obtain the best classification accuracy results in comparison with other models such as Inception, DenseNet etc. \cite{DBLP:journals/ijgi/KontopoulosMT21,makris2020covid}.

\section{Experimental Evaluation}
\label{exp_evaluation}

\subsection{Dataset Description}

\label{datasets_overview}

In general, trajectories can be classified into four major categories regarding mobility: people, vehicles, animals and natural phenomena \cite{sun2016overview}. 
The datasets employed for the evaluation of the different trajectory classification techniques, were selected according to some criteria:
\begin{enumerate}
\item the data are categorized in classes, so they can be used to evaluate classification problems.
\item they contain trajectories of different objects (people, vehicles, hurricanes, and animals). 
\item they present variations in their sampling rates, ranging from highly sampled to very low sampling rate.
\end{enumerate}

Many of these datasets have already being used in \cite{ferrero2018movelets} and \cite{makris2021evaluating} for comparing several classification methods.
A detailed description of the trajectory datasets is presented below. 

\subsubsection{Microsoft GeoLife dataset}

This GPS trajectory dataset \cite{zheng2010geolife,zheng2008understanding,zheng2009mining}, was collected in (Microsoft Research Asia) Geolife project by 182 users in a period of over three years (from April 2007 to August 2012). This dataset recorded a broad range of users’ outdoor movements, including not only life routines like go home and go to work but also some entertainments and sports activities, such as shopping, sightseeing, dining, hiking, and cycling. Furthermore, it contains several transportation modes as cars, bike, bus, and foot. As this dataset has a high sampling rate, some issues such as outliers and duplicated records needed to be fixed before applying data mining techniques. The pre-processing phase contains the following steps: a) removing duplicated records, b) splitting the trajectories where two consecutive points had more than 300 seconds of difference between them, as it represents a large gap in such dense dataset, c) removing trajectories with less than 100 points, as it represents a very small portion of time due to the density, d) excluding the transportation modes with too less trajectory examples, which are the classes of airplane, boat, running and motorcycle e) removing trajectories with unreal average velocity given the transportation mode, as for instance, trajectories labeled with walk with average velocity of more than 10m/s, and f) from the remaining trajectories, a proportion of 20\% of the trajectories of each transportation mode was selected, as some techniques were unable to perform in a reasonable amount time.

\subsubsection{Hurricane Dataset}
This dataset\footnote{\url{https://www.nhc.noaa.gov/data/}} is provided by the National Hurricane Service (NHS) and is related to Atlantic Hurricanes Database collected between 1950 and 2008. It contains sparse points with hurricane trajectories, where the hurricanes are classified in strength level. It has a low sampling rate and the class is the degree of the hurricane. 

\subsubsection{Animals Dataset}

This dataset\footnote{\url{https://www.fs.fed.us/pnw/starkey/index.shtml}} contains information of three animal species from the Starkey project. It contains sparse points of major habitat variables derived for radiotelemetry studies of elk, mule deer and cattle at the Starkey Experimental Forest and Range, 28 miles southwest of La Grande, Oregon. It has a low sampling rate and the class is the type of animal.

Table \ref{tab:overview_dataset} provides information about the datasets employed for evaluating the various trajectory classification methods.

\begin{table}
\centering
\label{tab:overview_dataset} 
\begin{tabular}{llllll}
\hline\noalign{\smallskip}
Dataset & Trajectories & Points & Sampling rate & Type \\ 
\noalign{\smallskip}\hline\noalign{\smallskip}
GeoLife & 1,763 & 953,966  & high & dense \\
Hurricanes & 1003 & 26,783  & low & sparse \\
Animals & 253 & 287,136 & low & sparse \\
\noalign{\smallskip}\hline
\end{tabular}
\caption{Datasets Overview}
\end{table}

\subsection{Trajectory Classification Techniques}

Trajectory classification techniques aim to identify the classes of the trajectories by extracting features that are capable of distinguishing the classes. In this research work, the proposed deep learning method is compared against four more methods. These methods, which are used for extracting trajectory features, were selected because of their best classification accuracy. Table \ref{tab:classification_techniques} summarizes the classification methods, the trajectory dimensions they support and the features they extract from trajectories that are used as input for the classification algorithms.

\begin{table}[h!]
    \centering
    \begin{tabular}{c|c|c}
        \textbf{Technique} & \textbf{Trajetory Dimensions} & \textbf{Features Extracted} \\
         \hline
         \multirow{4}{10em}{Dodge \cite{dodge2009revealing},  Zheng \cite{zheng2008understanding} and Xiao \cite{xiao2017identifying}} & \multirow{4}{10em}{Space and time}  & \multirow{4}{10em}{Numerical features, e.g. speed, acceleration, deviation rate, etc.
         }  \\
         & & \\
         & & \\
         & & \\
         \hline
         \multirow{4}{10em}{Movelets \cite{ferrero2018movelets}} & \multirow{4}{10em}{Space and Time} & \multirow{4}{10em}{Movelets extracted from space dimension + global features as indicated in \cite{ferrero2018movelets}
         } \\
         & & \\
         & & \\
         & & \\
         \hline
         \multirow{4}{10em}{TraClets} &
         \multirow{4}{10em}{Space and Time} & \multirow{4}{10em}{Image representations extracted from raw trajectories
         } \\
         & & \\
         & & \\
         & & \\
         \hline
    \end{tabular}
    \caption{Summary of the classification techniques, with the trajectory dimensions they use, and the features they extract from trajectories.}
    \label{tab:classification_techniques}
\end{table}

The techniques of Dodge \cite{dodge2009revealing}, Zheng \cite{zheng2008understanding} and Xiao \cite{xiao2017identifying} extract the numerical features from trajectories, e.g. the average speed, the maximum acceleration, etc., which are formulas that need the spatio-temporal dimensions from trajectories.
The Movelets \cite{ferrero2018movelets} extract the movelets, which are the sub-trajectories that better discriminate the classes.
Although the Movelets can deal with any trajectory dimension, Ferrero in \cite{ferrero2018movelets} indicated that the best dimension is the space, with the aggregation of some numerical features extracted from the whole trajectories. In TraClets, image representations are extracted from raw trajectory data.

\subsection{Experimental Results}

After extracting  the features, the Random Forest (RF) is trained and used to compare Dodge, Zheng, Xiao and Movelets, as it is commonly used by classification techniques in the literature. The works of \cite{ferrero2018movelets}, \cite{silva2019} and \cite{kontopoulos2021computer} have compared several classifiers including SVM (Support Vector Machine), MLP (Multi Layer Perceptron) and Random Forest, and they have shown a similar accuracy. We have selected RF because it performs faster than SVM and MLP.
Random Forest was constructed with $100$ decision trees with the default structure. 
The threshold values used by Dodge, Zheng and Xiao are those reported in the original paper. To extract the local features of \cite{dodge2009revealing} the sinuosity threshold is necessary to identify similar subtrajectories, and this threshold is given by the mean between the minimum and maximum sinuosity value.
To extract the Zheng features that require thresholds (e.g. the heading change rate) we used the method suggested in \cite{zheng2008understanding} for finding the thresholds, which consists on evaluating each feature individually for finding the best set of thresholds for each dataset. 
In the Movelets technique the Euclidean distance measure is used.
The datasets were split in a hold-out manner, in a proportion of $70\%$ for training and $30\%$ for testing.

In order to evaluate the classification performance, the Accuracy (ACC) metric was adopted. $Accuracy$ indicates how well a classification algorithm can discriminate the classes of the trajectories in the test set. As shown in Equation (\ref{accuracy_eq}), $ACC$ can be defined as the proportion of the predicted correct labels (true positives/negatives) to the total number of labels ($N$).

\begin{equation} \label{accuracy_eq}
Accuracy (ACC) = \frac{|Correctly\ Labeled\ Examples|}{N}
\end{equation}

The classification performance of VGG16 was evaluated on the effectiveness of trajectory pattern recognition. The datasets were split in the same hold-out manner as previous ($70\%$ - $30\%$) and reported the macro-average results. The Keras package and a TensorFlow backend were employed along with the Python programming language for training the deep learning models. Keras is a simple-to-use neural network library built on top of Theano or TensorFlow.

\begin{table}[h!]
    \centering
    \begin{tabular}{c|c|c|c|c|c}
        \textbf{Dataset} & \textbf{\cite{dodge2009revealing}} & \textbf{\cite{zheng2008understanding}} &  \textbf{\cite{xiao2017identifying}} & \textbf{\cite{ferrero2018movelets}} & \textbf{TraClets} \\
         \hline
         GeoLife & $81.01\%$ & $82.33\%$ & $83.77\%$ & $83.98\%$ & $\textbf{86.5}\%$ \\
         Hurricanes & $60.52\%$ & $58.22\%$ & $60.19\%$ & $\textbf{61.84}\%$ & $59.2\%$ \\
         Animals & $82.05\%$ & $87.17\%$ & $85.89\%$ & $91.02\%$ & $\textbf{92.15}\%$ \\
    \end{tabular}
    \caption{Summary of the classification results in terms of accuracy.}
    \label{tab:classification_results}
\end{table}

Table \ref{tab:classification_results} illustrates the classification accuracy of all classifiers over all datasets. It is shown that TraClets achieved the best accuracy in two out of three datasets with an accuracy of $86.5\%$ in the GeoLife dataset and $92.15\%$ in the Animals dataset while Movelets are the next best classifier succeeding TraClets with an accuracy of $83.98\%$ and $91.02\%$ respectively. On the other hand, Movelets achieved the best accuracy in the Hurricanes dataset ($61.84\%$). Furthermore, it is worth noting that the Hurricanes dataset challenged every classifier yielding the lowest classification accuracies.

\begin{figure}
\centering
\subfigure[GeoLife dataset.]{\label{fig:accloss_geolife}\includegraphics[scale=0.4]{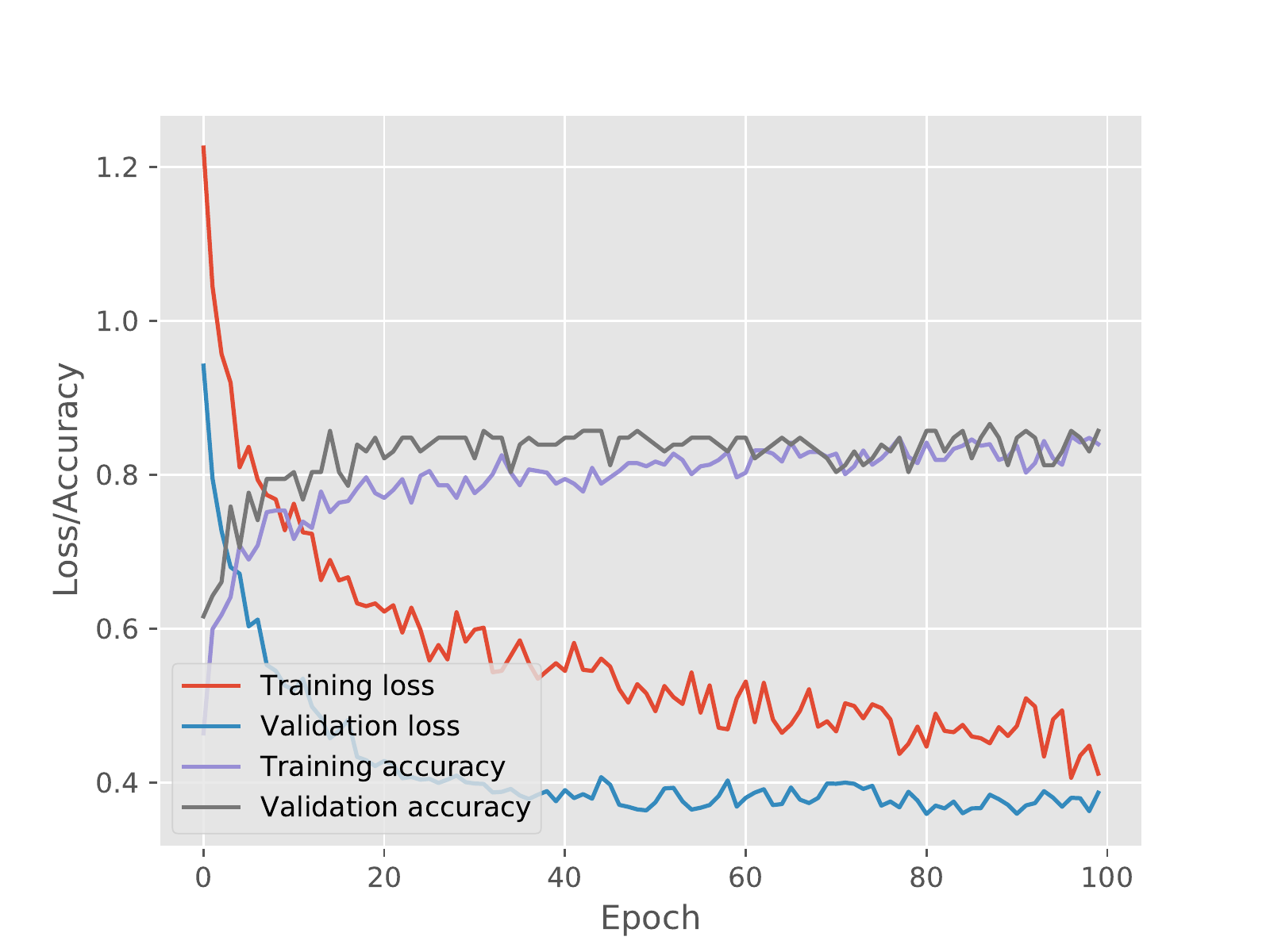}}
\subfigure[Hurricanes dataset.]{\label{fig:accloss_hurricanes}\includegraphics[scale=0.4]{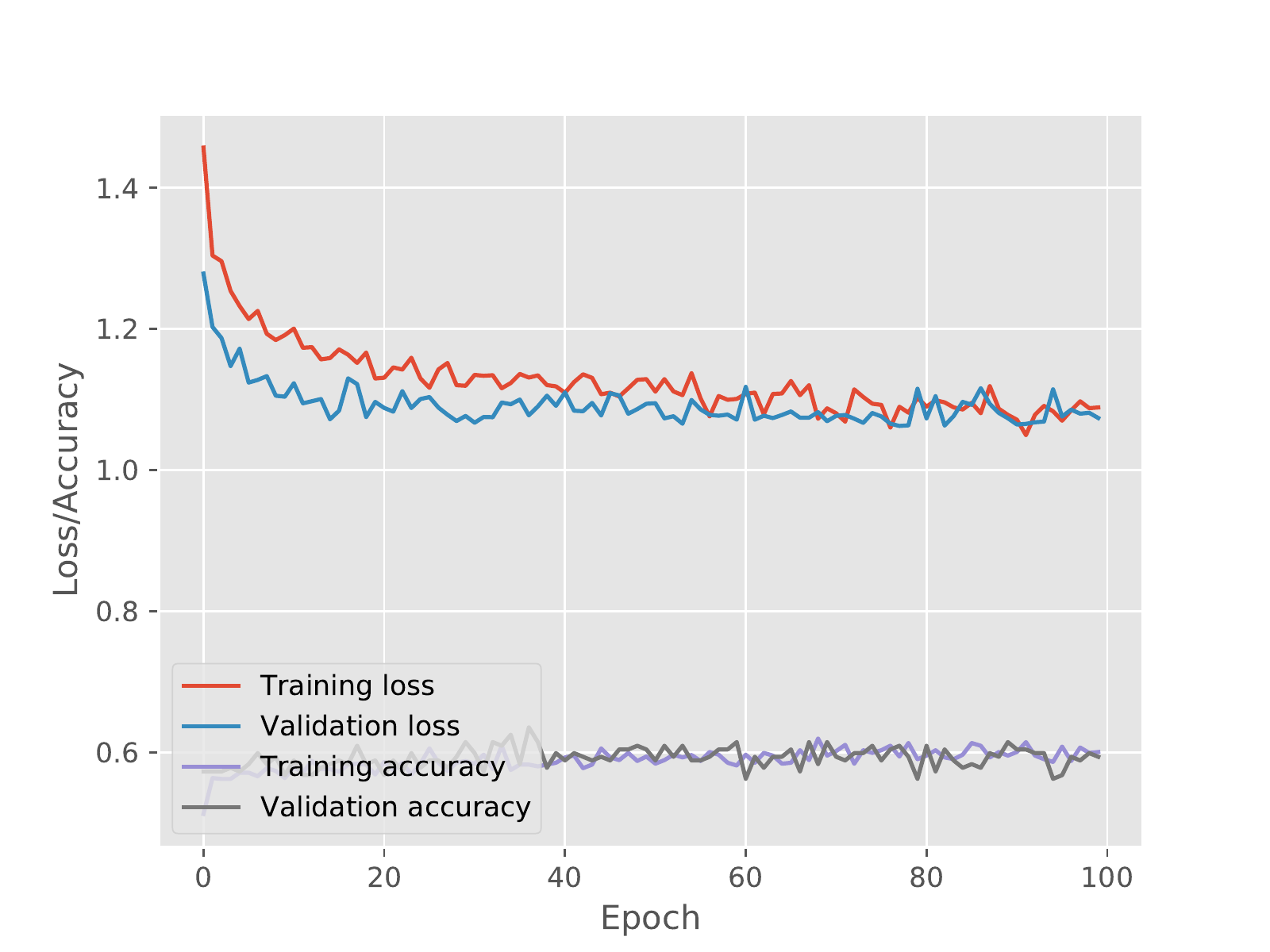}}
\subfigure[Animals dataset.]{\label{fig:accloss_animals}\includegraphics[scale=0.4]{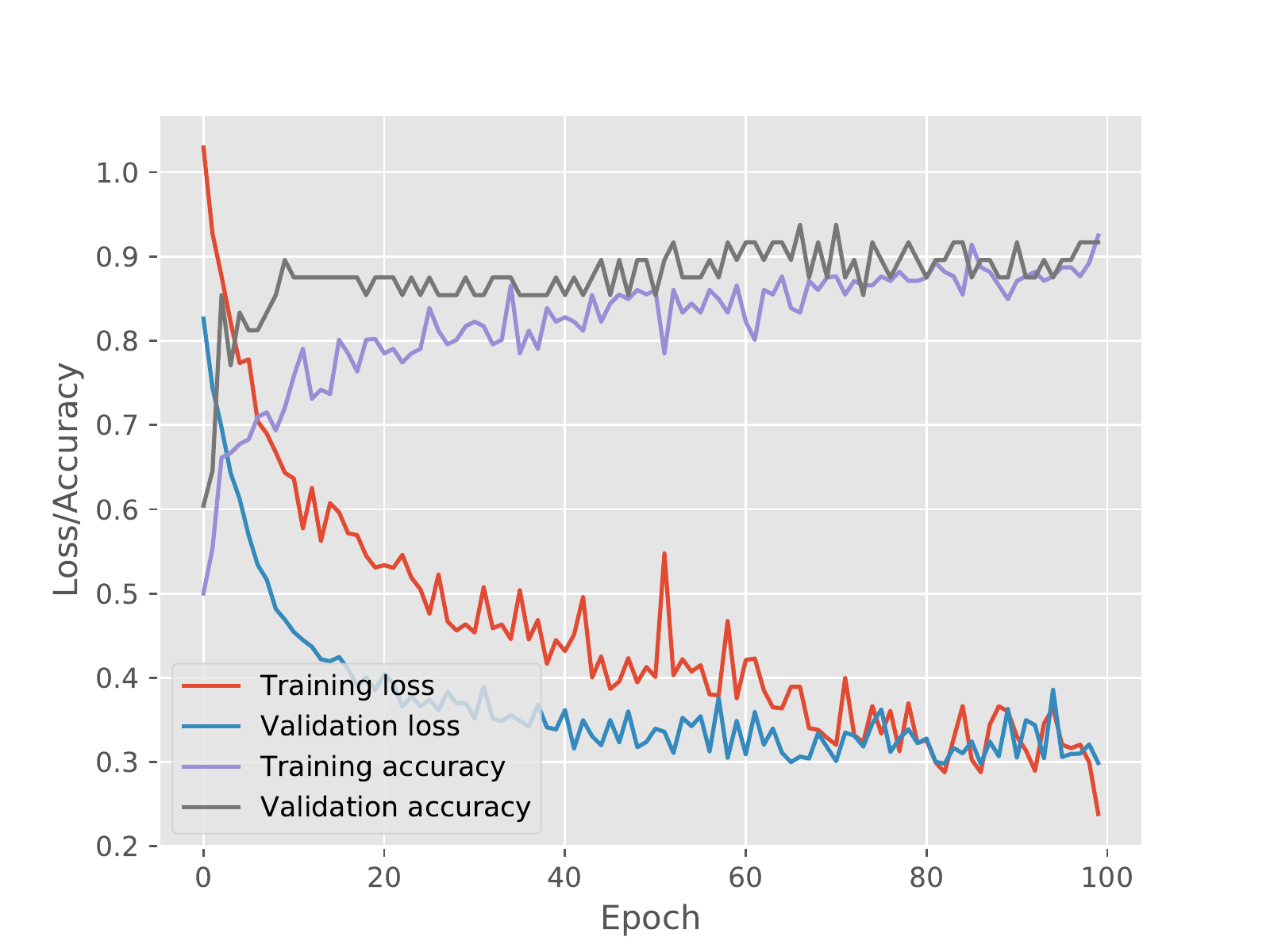}}
\caption{Accuracy and Loss during training.}
\label{traclets_results}
\end{figure}

To further visualize the TraClets' performance we illustrated the loss and the accuracy during the training process in Figure \ref{tab:classification_results}. TraClets demonstrates a smooth training process for all datasets, during which the loss gradually decreases and the accuracy increases. Moreover, it can be observed that the accuracy of both training and validation do not deviate much from one another, a phenomenon that can also be observed for the training and validation loss, indicating that the deep learning model does not overfit. Finally, in Figure \ref{fig:accloss_hurricanes} that demonstrates the training process in the hurricanes dataset, it can be seen that the accuracy does not increase much and the loss stops decreasing after epoch 20, indicating that the deep learning model cannot distinguish between the classes beyond that point.

The fact that the model cannot distinguish between the classes of hurricanes is also apparent in Table \ref{tab:traclets_results} that illustrates the precision, recall and f1-score of our approach in all datasets. It can be observed that the scores achieved in the hurricanes dataset are two low (F1-score of $0.26$). On the other hand, the classes of the rest of the datasets can be visually distinct from each other and for that reason the proposed approach achieved an F1-score of $0.83$ and $0.88$ in the GeoLife and the Animals dataset respectively. It is worth noting that the proposed methodology for trajectory classification achieved a state-of-the-art classification performance with no configurations, i.e., no manual adjustment of the speed and acceleration increments was performed as mentioned in Sections \ref{speed_rep} and \ref{accel_rep} respectively. Therefore, we can safely assume that TraClets can act as a universal trajectory classification technique that can work well in a wide range of datasets requiring minimal or no configurations in most cases to achieve a high-accuracy classification performance.

\begin{table}[h!]
    \centering
    \begin{tabular}{c|c|c|c}
        \textbf{Dataset} & \textbf{Precision} & \textbf{Recall} &  \textbf{F1-score} \\
         \hline
         GeoLife & $0.84$ & $0.83$ & $0.83$ \\
         Hurricanes & $0.26$ & $0.28$ & $0.26$ \\
         Animals & $0.94$ & $0.86$ & $0.88$ \\
    \end{tabular}
    \caption{Summary of the classification results of TraClets.}
    \label{tab:traclets_results}
\end{table}

\section{Conslusion}
\label{conclusion}

This work presented a novel and high-accuracy trajectory classification approach in an attempt to provide an efficient and alternative way to treat the problem of trajectory classification. The proposed methodology employs a state-of-the-art deep learning algorithm and creates a universal approach for the classification of trajectories. This universal approach can be achieved through image representations of trajectories, called TraClets, that amplify the distinct visual difference that most of the moving objects' trajectories, if not all, have in three distinct domains.

To demonstrate the applicability of our work, three datasets were used to evaluate the performance of the proposed approach. As deep learning approaches require a large amount of data in order to perform an accurate classification, transfer learning was also employed. Experimental evaluation demonstrated that TraClets achieve a state-of-the-art classification accuracy that is comparable to or better than other state-of-the-art trajectory classification algorithms in the literature.

As a future work, we plan on extending and evaluating our approach on more domains such as airplanes that may require to take into account the height instead of only longitude and latitude.



\section*{Acknowledgments}

This work was supported by the MASTER and SmartShip Projects through the European Union's Horizon 2020 research and innovation program under Marie-Sklodowska Curie Grant Agreement Nos. 777695 and 823916, respectively. The work reflects only the authors' view, and the EU Agency is not responsible for any use that may be made of the information it contains. The works was also supported by the Brazilian agencies CAPES PRINT, CNPQ and FAPESC.





\end{document}